\documentclass[journal]{IEEEtran}
\usepackage{algorithm}
\usepackage{algcompatible}
\usepackage[ampersand]{easylist}
\usepackage{algorithm}
\usepackage{pbox}
\usepackage{amsfonts,amssymb}
\usepackage{amsmath}
\usepackage{changepage}
\usepackage{tikz}
\usepackage{soul}
\usepackage{multirow}
\usepackage{booktabs}
\usepackage{threeparttable}
\usepackage{tabularx}
\usepackage{hyperref}
\hypersetup{colorlinks,linkcolor={blue},citecolor={blue},urlcolor={blue}}
\usepackage{url}

\usepackage{pbox}
\usepackage{makecell}
\usepackage{notoccite}
\usepackage{bm}
\newsavebox{\tablebox}
\usepackage{array}
\newcolumntype{L}[1]{>{\raggedright\arraybackslash}p{#1}}

\begin{document}
\title{A Reference Architecture for Plausible Threat Image Projection (TIP) Within 3D X-ray Computed Tomography Volumes}

\author{Qian Wang~\IEEEmembership{Member,~IEEE}, \and
        Najla Megherbi, \and
        Toby P. Breckon~\IEEEmembership{Member,~IEEE}% <-this % stops a space
\IEEEcompsocitemizethanks{\IEEEcompsocthanksitem Qian Wang is with Department of Computer Science, Durham University, UK. E-mail: qian.wang173@hotmail.com; 
}% <-this % stops an unwanted space
\thanks{Najla Megherbi was with School of Engineering, Cranfield University, UK. E-mail:
najla.megherbi@gmail.com;}
\thanks{Toby P. Breckon is with Department of Engineering and Department of Computer Science, Durham University, UK. E-mail:toby.breckon@durham.ac.uk.}
}
\markboth{}%
{Shell \MakeLowercase{\textit{et al.}}: Bare Demo of IEEEtran.cls for IEEE Journals}

\maketitle

\begin{abstract}
Threat Image Projection (TIP) is a technique used in X-ray security baggage screening systems that superimposes a threat object signature onto a benign X-ray baggage image in a plausible and realistic manner. It has been shown to be highly effective in evaluating the ongoing performance of human operators, improving their vigilance and performance on threat detection. However, with the increasing use of 3D Computed Tomography (CT) in aviation security for both hold and cabin baggage screening a significant challenge arises in extending TIP to 3D CT volumes due to the difficulty in 3D CT volume segmentation and the proper insertion location determination. In this paper, we present an approach for 3D TIP in CT volumes targeting realistic and plausible threat object insertion within 3D CT baggage images. The proposed approach consists of dual threat (source) and baggage (target) volume segmentation, particle swarm optimisation based insertion determination and metal artefact generation. In addition, we propose a TIP quality score metric to evaluate the quality of generated TIP volumes. Qualitative evaluations on real 3D CT baggage imagery show that our approach is able to generate realistic and plausible TIP which are indiscernible from real CT volumes and the TIP quality scores are consistent with human evaluations.
\end{abstract}


% \fi
\IEEEpeerreviewmaketitle

\section{Introduction}\label{sec:introduction}
% Why/What/How?
Scanning passenger baggage using X-ray technology is a mandatory process in airports and other public transportation for security. Although automatic threat material and prohibited item detection using advanced machine learning techniques have been studied \cite{akcay2018using,akcay18ganomaly,wang2019approach}, they have not yet achieved maturity whereby human operators can be completely replaced. As such the performance of human operators can vary depending on experience, fatigue and baggage item complexity.
Threat Image Projection (TIP) is a technique applied in X-ray image based baggage screening systems to monitor the ongoing performance of human operators. TIP is used to generate plausible and realistic X-ray baggage images containing threat signatures (e.g., firearms, improvised explosive devices, etc.) by projecting fictional threat object images onto X-ray images of real passenger bags present within the live aviation security process. It is a little known fact that TIP is a legally mandated process by both national and international aviation security regulations \cite{otac178-3,tip-uk}. 

% The benefits of TIP 
Using TIP in X-ray security scanners has been shown to be effective in improving the vigilance and attention of human operators, hence improving the overall performance of threat detection \cite{hofer2005using,cutler2009use}. The benefits of TIP systems are multi-fold.
TIP systems make it possible for operators to encounter baggage images with threat objects more frequently during their regular working patterns by randomly applying TIP to benign passenger bags so that they can get familiar with potential real yet rare threats in order to improve their detection ability \cite{cutler2009use}. Research also suggests operators are motivated and more attentive to do well when knowing that TIP systems are enabled and their performance is monitored \cite{otac178-3}. In addition, TIP systems record the performance of individual operators such that his information could be further analysed and used to customize training plans.

These benefits, however, are only achievable if TIP systems are properly used and managed \cite{cutler2009use}. For example, it is important how frequently an operator should be exposed to TIP during their work pattern. On the other hand, the management of TIP library is critical to the effectiveness of TIP systems. According to \cite{otac178-3}, 
the TIP library shall contain a minimum of 1,000 virtual images and 250 threat objects captured in different orientations. The TIP library needs to be updated each year with no fewer than 100 virtual images replaced with new ones. To satisfy these requirements, an effective algorithm of plausible and realistic TIP image generation is crucial.

% challenges in 3D TIP
Recently, 3D Computed Tomography (CT) scanners have seen increasing deployment in airports for baggage screening \cite{3dscanners2017}. A recent study in \cite{hattenschwiler2019detecting} shows the superior performance of threat detection using 3D CT imaging against traditional 2D X-ray images. However, 3D TIP within CT volumes is still a very challenging problem due to a number of additional factors. Firstly, some form of 3D CT volume segmentation is required to both isolate the bounds of the 3D threat object (source) and the exterior boundary and internal void regions of the 3D baggage item (target). Most CT volume segmentation algorithms are targeted at medical images \cite{litjens2017survey} which can not be readily applied to baggage volumes \cite{megherbi13segmentation}. Secondly, inserted threat objects have to avoid intersection with existing items in the target baggage volume and additionally exhibit artefacts consistency with those of the original scanned objects already present therein. \cite{megherbi2012fully}. 

% Our solution
To address aforementioned issues in 3D TIP, we extend the work in \cite{megherbi2012fully} and present a novel approach for fully automatic threat image projection within 3D CT security imagery. Our approach consists of four components: threat isolation, void determination, object insertion optimisation and metal artefact generation. Specifically, a threat volume is segmented into the background, threat body and uncertain regions (threat isolation), whilst a baggage volume is segmented into background, inner-void and bag-content regions (void determination). The segmentation results are used to evaluate the quality of a given insertion location and orientation. The optimal insertion is derived by particle swarm optimisation (object insertion optimisation). Finally, metal artefact generation \cite{megherbi2013radon} is applied to the generated TIP to enhance the plausibility. In summary, the paper has the following contributions:
\begin{itemize}
    \item[-] it is among the first attempts to address the threat image projection in 3D CT volumes to our best knowledge;
    \item[-] the proposed framework integrates 3D object segmentation and particle swarm optimisation algorithms towards the generation of realistic and plausible threat image projection;
    \item[-] the proposed approach has been validated on real baggage data collected from airports and the experimental results demonstrate its effectiveness from both qualitative and quantitative perspectives.
\end{itemize}
% Contribution
% Structure of the paper
The remainder of this paper is structured as follows: prior works relevant to 2D and 3D TIP are reviewed in Section \ref{sec:relatedwork}; we present details of our approach for 3D TIP in Section \ref{sec:3dtip}; qualitative evaluations of each component in our approach are given in Section \ref{sec:experiment}; finally, we discuss limitations existing in the current approach and potential directions of future work in Section \ref{sec:discussion} and conclude the paper in Section \ref{sec:conclusion}.

\section{Related Work}\label{sec:relatedwork}
In this section, we make a thorough review of TIP related works including those focusing on 2D and 3D imagery.
\subsection{2D TIP}\label{sec:2dtip}
The concept of threat image projection in X-ray imagery dates back to 1990s towards the enhancement of X-ray baggage screening performance in airports \cite{nadler1994airport, fobes1996operational}. As TIP within 2D imagery is essentially to superimpose a threat X-ray image onto a baggage X-ray image in a random position, it is a relatively simple technique in terms of contemporary image processing. As a result, literature on 2D TIP mainly focused on the system design \cite{neiderman2005threat,gudmundson2011method} or performance evaluation \cite{hofer2005using,cutler2009use,nadler1994airport, fobes1996operational} rather than image processing details. For example, Neiderman et al. \cite{neiderman2005threat} designed and patented a means for training and testing baggage screening operators using the 2D TIP technique. 
An exception is \cite{gudmundson2011method} in which the authors presented the details of combining distorted threat images with baggage images to generate realistic and diverse TIP. In addition, our recent work \cite{bhowmik2019good}, focusing on the investigation of TIP based data augmentation for object detection in X-ray baggage images, also provides some details of viable 2D TIP gleaned from various obfuscated sources \cite{neiderman2005threat,gudmundson2011method,rogers2016threat}.
%In contrast to TIP in 2D images, 3D TIP is a much more complex and challenging task of generating realistic and plausible 3D volumes.

Except for baggage screening, 2D TIP was also applied in video surveillance  \cite{neil2007threat,donald2015cctv} and cargo screening systems \cite{rogers2016threat}. Neil et al. \cite{neil2007threat} discussed the challenge and plausibility of applying TIP in video surveillance. Donald et al. \cite{donald2008vigilance} further discussed how TIP (or IGO, Inserted Graphic Objects) could improve vigilance performance, target incident detection rate, and design considerations for TIP images in video surveillance. Quantitative evaluations conducted by Donald \cite{donald2015cctv}, however, disclosed IGO were not effective in enhancing the detection of significant events in video surveillance. The reason, as discussed by the author, could be multifold and applying TIP in video streams is quite different from doing it in X-ray images. 

Rogers et al. \cite{rogers2016threat} applied X-ray TIP techniques in cargo screening tasks. They proposed a framework extracting threat masks from X-ray images and projecting them onto benign X-ray images to generate realistic TIP. Quantitative evaluations indicated the generated TIP and real X-ray images containing threats were indistinguishable.  In addition, transformations were made to inject variation into the threat signatures to generate a very large number of realistic TIP data for training deep learning based object detection algorithms. Among seven types of transformations employed in \cite{rogers2016threat}, threat insertion position and rotation are also used in our approach. The employment of such transformation in our approach aims to improve the plausibility of generated TIP which is not a problem in 2D TIP, while in \cite{rogers2016threat}, the transformation aims to diversify generated TIP for data augmentation. 

\subsection{3D TIP}\label{sec:3dtip-related}
Early attempts were made to extend 2D X-ray TIP to 3D TIP but unsuccessful due to obvious visual imaging artifacts which could provide cues for scanner operators to readily recognize the presence of TIP \cite{yildiz20083d}. To address this issue, Yildiz et al. \cite{yildiz20083d} proposed to project threat objects into the sinogram space instead of the original imagery space. They claimed plausible TIP could be generated but their insertion positions were manually decided and the evaluations were conducted on uncluttered image examples without explicit metal artefact generation. Megherbi et al. \cite{megherbi2012fully} proposed an approach to fully automatic 3D TIP and applied it to densely cluttered 3D CT baggage volumes (which is subsequently identified as prior work in the commercial implementations of \cite{chen2017ct,durzinsky2018}). The approach consisted of three main components: void determination, object insertion location determination and metal artefacts generation. The work presented in this paper is based on \cite{megherbi2012fully} but with notable variations and improvement within the stages of threat isolation and object insertion.

\section{Automatic 3D Threat Image Projection} \label{sec:3dtip}
\begin{figure*}[t]
    \centering
    \includegraphics[width=\textwidth]{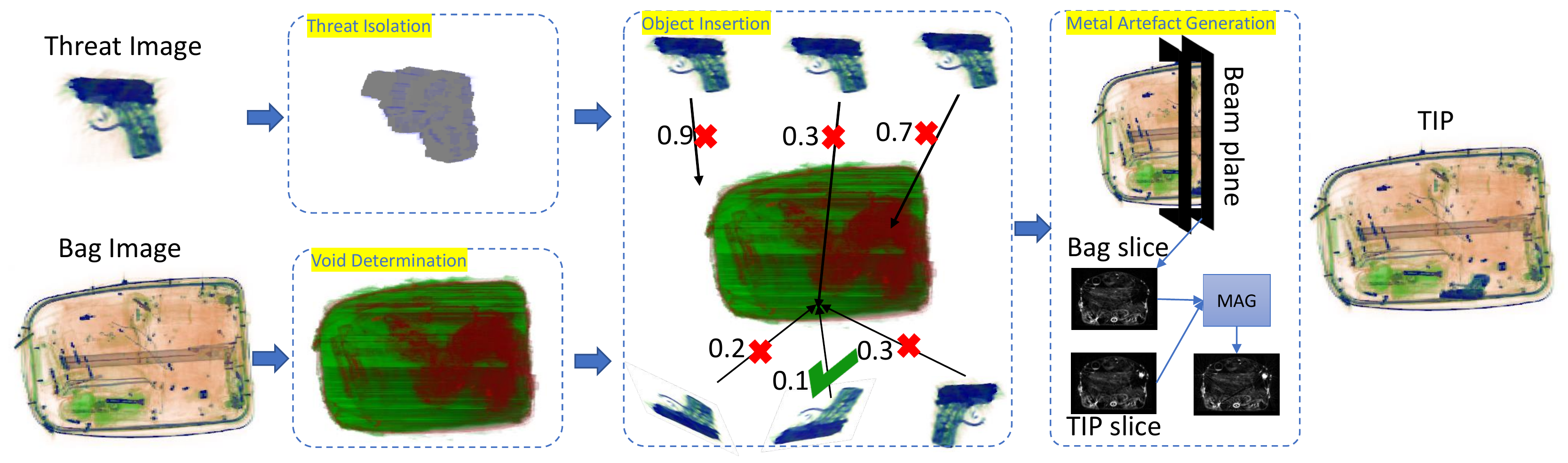}
    \caption[framework]{The framework of proposed 3D threat image projection approach; given a threat CT volume and a baggage CT volume as inputs, a plausible TIP is generated as the output of the approach with the pipeline consisting of four components: threat isolation, void determination, object insertion optimisation and metal artefact generation.}
    \label{fig:framework}
\end{figure*}
Our approach to 3D threat image projection is composed of four parts: threat isolation, void determination, object insertion optimisation and metal artefact generation. The framework of our approach is illustrated in Figure \ref{fig:framework}. 

% threat image segmentation overview
Threat isolation aims to segment threat objects from the background in threat volumes $\mathcal{I}^{thr}$. These threat objects of interest are prepared beforehand and scanned in a controlled condition (e.g. background voxels with lower values than threat object voxels) for easy segmentation. The subsequent thresholding and morphological operations used in our approach will be described in the following subsection.

% bag image segmentation overview
Void determination aims to segment baggage CT volumes $\mathcal{I}^{bag}$ into three regions: outer region, inner void and bag content. Different costs will be incurred when the threat object is projected into these three regions. As a result, bag volume segmentation results in a projection cost map of the bag volume indicating the cost of voxels onto which the threat object is projected.
% insertion
With the segmented threat object and the projection cost map of a bag volume, the insertion is boiled down to an optimisation problem which aims to find optimal insertion locations and threat object orientations. Particle swarm optimisation \cite{kennedy2010particle} is used in our approach as one of the enabling techniques.
% MAG
To enhance the plausibility of the generated TIP volume, we apply metal artefact generation \cite{megherbi2013radon} as post processing to generate plausible metal artefacts in the TIP volume.

In the following subsections, we will present four parts of our threat image projection approach in more detail. 

\subsection{Threat Isolation}\label{sec:threatSeg}
A variety of threat objects including firearms and improvised explosive devices could be used to generate TIP. To make segmentation easy and accurate, we assume the threat objects are scanned in controlled conditions. Every threat object will be scanned individually with only low-density supporting objects (e.g., foam) if necessary. As a result, threat object volumes (source) are almost free of noise except when metal components exist in the threat objects themselves. Special care needs to be taken to get rid of any residual artefact noise surrounding the body of the threat object.

\begin{figure}[t]
    \centering
    \includegraphics[width=0.45\textwidth]{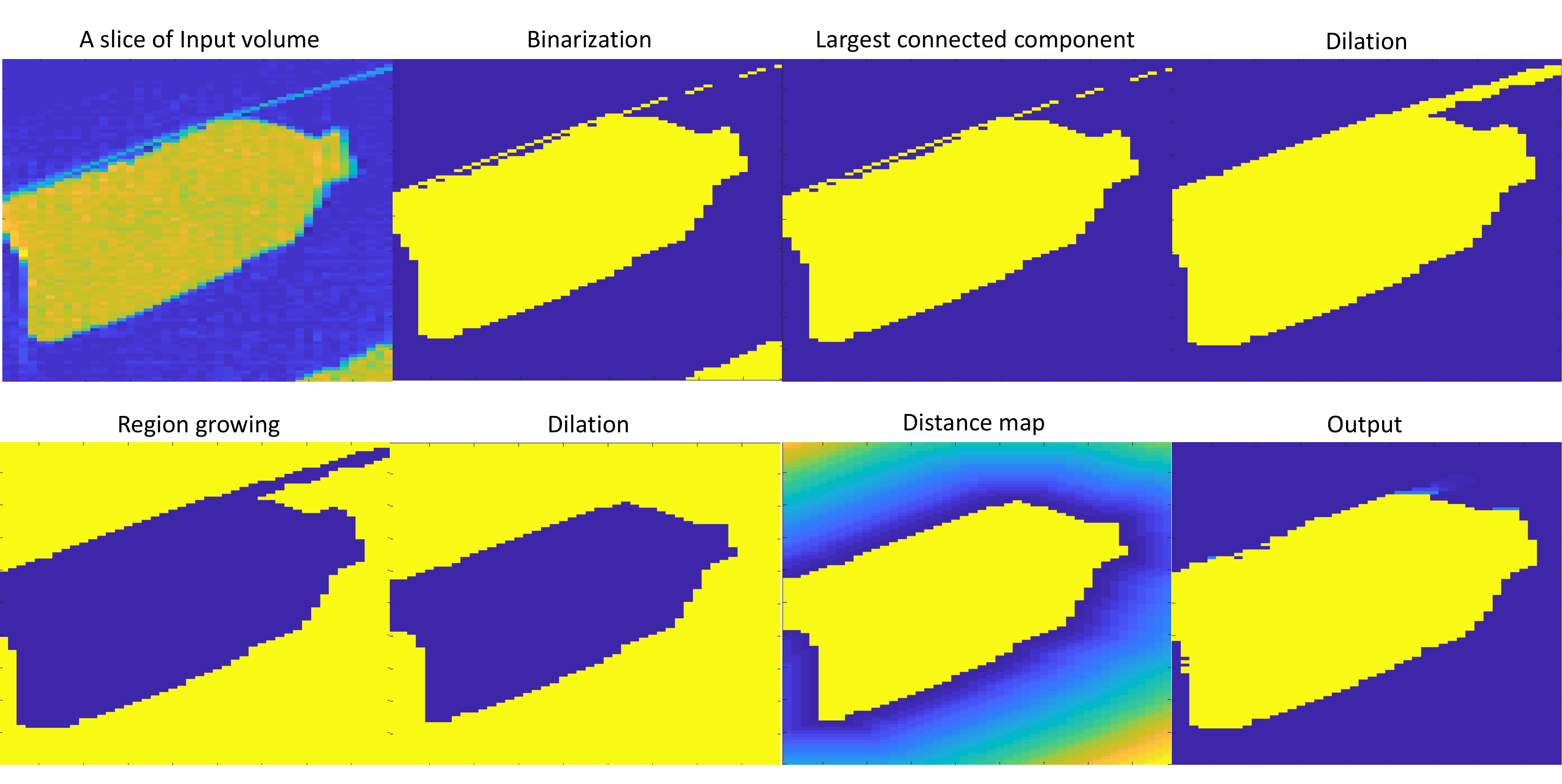}
    \caption[framework]{The pipeline of threat isolation algorithm; slices of a CT volume of a source item bottle is used here for illustration while the segmentation algorithm is actually applied to 3D CT volumes which are segmented into three regions:  the threat body region in yellow, the background in blue and the uncertain region in a gradient colour.}
    \label{fig:threatSegAlg}
\end{figure}

The pipeline of our threat isolation is shown in Figure \ref{fig:threatSegAlg}. It takes a threat volume as the input and outputs a cropped 3D volume of the threat object which is ready to be inserted in a benign bag volume target for TIP generation. CT volumes contain noise with small non-zero voxel values in the void region. To remove the effects of such noise, in the first step of threat isolation, we set a threshold value to binarise the input CT volume. Whilst this simple thresholding process will isolate the threat object in most cases, there could be special cases where the threat objects have an internal sub-void which should be considered as a part of the threat object. In addition, there could exist more significant noise in the CT volume which can not be removed by such simple thresholding. To handle these special cases, we develop a robust threat isolation approach advancing upon that of \cite{megherbi2012fully}. 

Specifically, connected component labelling (CCL) \cite{rosenfeld1976ccl} is applied to the binary volume derived from the thresholding process. The resulting labelled connected components could belong to either the threat object or background noise. Due to the fact that the noise components have far fewer voxels than the threat object, we only reserve the largest labelled connected component as the segmented threat object. As such we have successfully removed the noise in the volume but the resulted threat object can still have an internal void. To ensure that internal voids are treated as a part of the threat object, we try to conversely determine the exterior boundary of the threat object. Subsequently, the threat object including the possible void space inside can be derived accordingly. For this purpose, we use a region growing method \cite{adams1994seeded} to segment the exterior region in a threat volume. The region growing seed is usually set to the upper-left upper-leftmost voxel such that this will not be a theat object voxel within a controlled condition CT image scan. To use region growing the threat boundary should be closed so that the region cannot mistakenly grow into the threat object. To this end, we apply morphological dilation to the isolated threat component derived from the connected component labelling in the previous step.

Region growing is able to segment the non-threat (background) region in the threat volume (source). To reduce the noise surrounding the threat body, a dilation operation is applied to the non-threat region. The dilation operation transforms the voxels close to the threat boundary into the background and effectively removes some noisy voxels from the threat object. However, it could also lead to damage to the true voxels of the threat object. To alleviate this issue, we consider the voxels removed by the background dilation form an uncertain region since the voxels of this region could belong to the threat object or be noise.

Now the threat volume is segmented into three parts: the threat object, the uncertain region and the background. A minimum 3D volume of the threat object is cropped and most of the background region is removed. To facilitate the presentation, we use $\bm{I}^{thr}$ to denote the cropped 3D volume of a threat object and a 3D indicator matrix $\bm{M}^{thr}$ is used to represent the segmentation results. The element value of $\bm{M}^{thr}(i,j,k)$ is determined as follows:
\begin{equation}
\label{eq:threat}
\bm{M}^{thr}(i,j,k) = \left \{
\begin{array}{ll}
1, & threat \quad object,\\
1/d^2_{ijk}, & uncertain \quad region,\\
0 ,&  background.
\end{array}
\right.
\end{equation} 
where $d_{ijk}$ is the distance of voxel $(i,j,k)$ to the boundary of the threat object, which can be calculated by distance transform method proposed by Maurer et al. \cite{maurer2003linear}. Equation (\ref{eq:threat}) results in a 3D volume composed of three different regions: threat body voxels indicated by ones, background voxels with zero values and uncertain voxels in the range of $0-1$. The indicator matrix $\bm{M}^{thr}$ will serve as a weight matrix to extract the threat object from the original CT volume for TIP generation. The uncertain voxels far from the threat object will have lower weights so that the sharp transition effect can be alleviated when inserting the threat into a benign bag volume. As a result, the resultant TIP look more plausible by using the indicator matrix in Eq. (\ref{eq:threat}) for initial threat isolation.

\subsection{Void Determination}\label{sec:bagSeg}
\begin{figure}[t]
    \centering
    \includegraphics[width=0.48\textwidth]{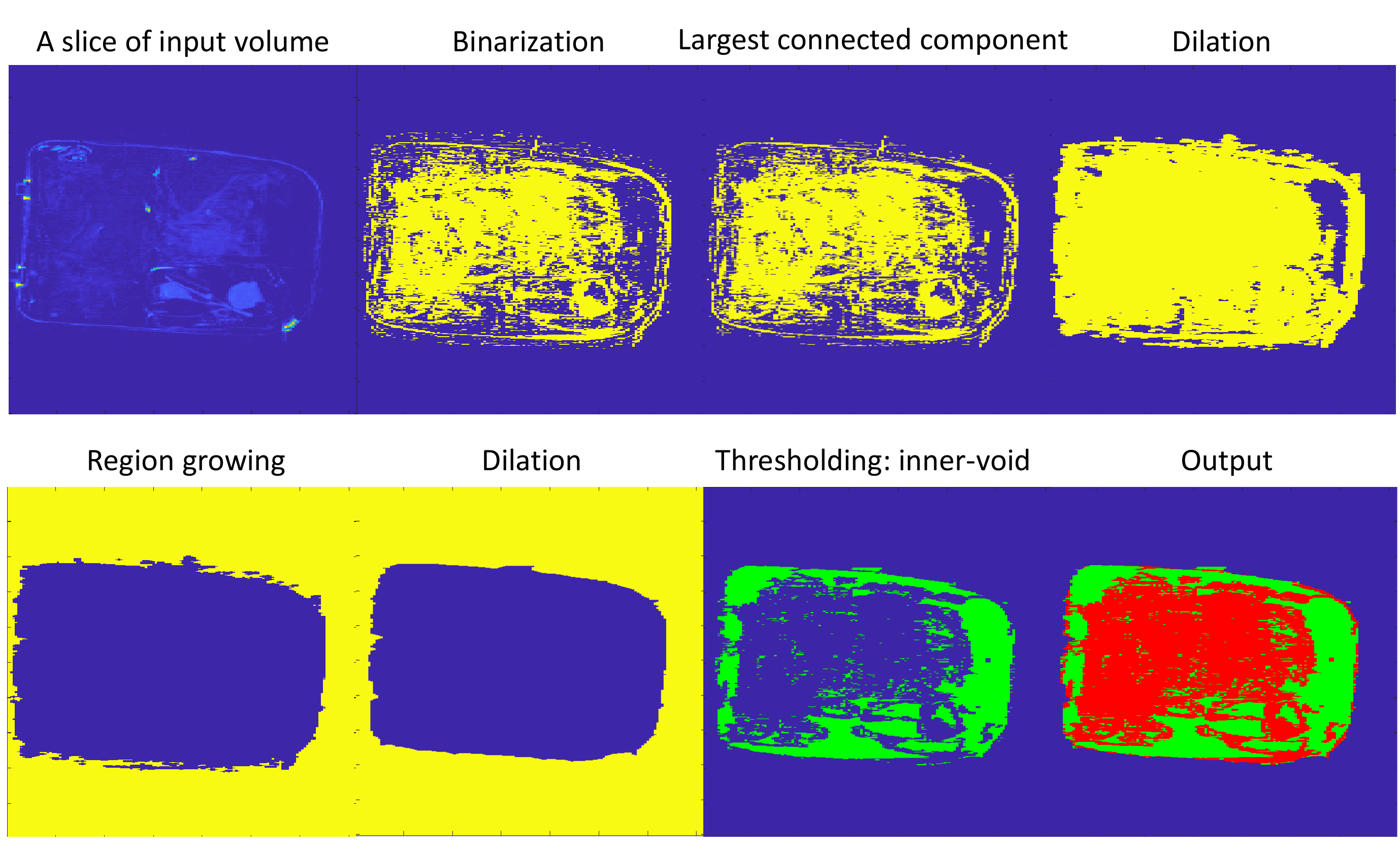}
    \caption[framework]{The pipeline of void determination algorithm (left to right, top to bottom); slices of a 3D CT volume of a suitcase are used for illustration while the algorithm is actually applied to 3D CT volumes which are segmented into three regions: the outer-bag region in blue, the inner-void region in green and the bag-content region in red.}
    \label{fig:bagSegAlg}
\end{figure}

To insert the segmented threat into a plausible location in the bag volume, we require to understand different regions in the bag volume. We propose a bag volume segmentation method to segment a bag volume into: \textit{outer-bag} (background), \textit{bag-content} and \textit{inner-void} regions. Similar to the threat volume segmentation, the pipeline of bag volume segmentation is composed of several morphological operations as illustrated in Figure \ref{fig:bagSegAlg}. The pipeline takes a 3D CT volume as input and outputs a indicator matrix representing the segmentation results.

A similar approach to threat volume segmentation is used here to segment the background region of bag volumes. The original CT volume is firstly binarised by thresholding. Subsequently, the largest connected component is extracted by 3D CCL as the volume of a bag which is further dilated to ensure the boundary is closed. The background is segmented by region growing from a random seed (usually set as the upper left most voxel) outside the volume of the bag. 

To segment a volume of bag into an \textit{inner-void} region and a \textit{bag-content} region, a simple thresholding is applied so that voxels of smaller values than the threshold form the \textit{inner-void} region and others form the \textit{bag-content} region. 

We use $\bm{I}^{bag}$ to denote the 3D volume matrix of a bag volume $\mathcal{I}^{bag}$ and subsequently the segmentation results can be represented by a 3D projection cost map $\bm{M}^{bag}$ as follows:
\begin{equation}
\label{eq:bag}
\bm{M}^{bag}(i,j,k) = \left \{
\begin{array}{ll}
0, & inner{\text-}void,\\
\tilde{v}_{ijk}, & bag{\text-}content,\\
c, &  outer{\text-}bag.
\end{array}
\right.
\end{equation} 
where $c$ is a large positive constant indicating that a big cost will be incurred if threat object is projected to the \textit{outer-bag} region, and $\tilde{v}_{ijk} = \bm{I}^{bag}(i,j,k)\times c/m$ denotes the cost of projecting onto \textit{bag-content} voxels which is equal to the normalized intensity value of voxel $(i,j,k)$ where $m$ is the maximum voxel density value in the CT volume.

\subsection{Object Insertion Optimization} \label{sec:insertion}
Inserting a segmented threat object into a benign bag in the 3D CT imagery involves the determination of optimal insertion locations and orientations. To enable plausible and realistic TIP, it is important to find suitable locations in a benign bag and proper orientations of the threat object. Ideally, we tend to insert a threat object into the \textit{inner-void} region in the bag volume. In cases where a large void region is available in the benign bag, we need to consider the effect of gravity and insert it into a lower position so that the inserted threat object will not appear to implausibly levitate unsupported. In practice, however, most baggage is cluttered without enough void space for big threat objects. In these cases we allow the threat object to be inserted into regions where voxel intensities are low. Regions of low voxel values are usually occupied by clothes and it is highly plausible to have a threat object concealed in clothes within baggage. In our TIP framework, we formulate an optimization problem to ensure optimal insertions and use Particle Swarm Optimization (PSO) \cite{kennedy2010particle} to solve the problem. Once the optimal location and orientation have been derived, a simple blending approach is employed to project the threat object into the bag volume and generate a final TIP volume output.
\subsubsection{Optimizing Insertion Location and Orientation}\label{sec:opt}
% describe the optimization problem: objective, parameters to optimize
Finding an optimal insertion location and orientation is essentially an optimization problem. The variables to optimize are denoted by $\bm{\Theta} = \{x,y,z,\alpha,\beta,\gamma\}$, where $x$, $y$, $z$ are the insertion coordinates within the volume of bag (target) and $\alpha$, $\beta$, $\gamma$ are the rotation angles of the threat object (source) within three planes ($yz$, $xz$ and $xy$). We use $\tilde{\bm{M}}^{thr}$ to denote the 3D volume matrix after rotating $\bm{M}^{thr}$ by angles of $\alpha, \beta, \gamma$ within three planes.
\begin{equation}
    \label{eq:rotation}
    \tilde{\bm{M}}^{thr} = f(\bm{M}^{thr}|\alpha,\beta,\gamma)
\end{equation}
where $f(\cdot)$ is a rotation function which could be implemented by spline interpolation \cite{shoemake1985animating, scipy-rotation}. We use $\tilde{\bm{M}}^{bag}$ to denote a 3D volume cropped from the projection cost map $M^{bag}$ of the bag volume. The cropping is conditioned on the coordinates $x,y,z$ and the size of rotated threat volume $\tilde{\bm{M}}^{thr}$:
\begin{equation}
    \label{eq:cropping}
    \tilde{\bm{M}}^{bag} = g(\bm{M}^{bag}|x,y,z,\Delta_x, \Delta_y, \Delta_z)
\end{equation}
where $g(\cdot)$ represents the cropping process, $\Delta_x, \Delta_y, \Delta_z$ are sizes of the cropped 3D volume and equal to those of the rotated threat volume $\tilde{\bm{M}}^{thr}$.

The objective of the optimization problem is formulated as follows:
\begin{equation}
\label{eq:obj}
\min_{\bm{\Theta}} ||\bm{M}^{tip}||_1 + \lambda_1 ||\tilde{\bm{M}}^{tip}||_1 + \lambda_2 y
\end{equation}
where
\begin{equation}
    \label{eq:tiphat}
    \tilde{\bm{M}}^{tip}(i,j,k) = H(\bm{M}^{tip}(i,j,k)-c')
\end{equation}
and
\begin{equation}
    \label{eq:mtip}
    \bm{M}^{tip} = \tilde{\bm{M}}^{thr} \odot \tilde{\bm{M}}^{bag}.
\end{equation}
$H(\cdot)$ in Eq. (\ref{eq:tiphat}) is a unit step function applied to all elements of $\bm{M}^{tip}$ so that $\tilde{\bm{M}}^{tip}(i,j,k) = 1$ when $\bm{M}^{tip}(i,j,k)$ is greater than the constant parameter $c'$, otherwise 0. The operator $\odot$ between two matrices in Eq. (\ref{eq:mtip}) denotes the Hadamard product. The operator $||\cdot||_1$ in Eq. (\ref{eq:obj}) is the entrywise matrix 1-norm which calculates the sum over the absolute values of all elements in a matrix; $\lambda_1$ and $\lambda_2$ are two hyper-parameters adjusting the weights of different terms in the objective function.

% a further explanation of the objective function
Minimizing the first term in Eq. (\ref{eq:obj}) ensures the threat object to be inserted in a region with the lowest average voxel intensity. However, it could result in a solution where small volumes of high-intensity voxels close to the low-intensity region are selected. Such regions may have the lowest average intensities but make the insertion less plausible. For example, the threat may be inserted into an empty corner of a bag but with a small part outside the bag. To address this issue, we have the second item in Eq. (\ref{eq:obj}) which aims to minimize the number of voxels whose values are greater than a threshold $c$ in the selected bag regions. The third item aims to limit the coordinate value in the direction of gravity so that it tends to insert the threat object in a lower location within the bag.

\subsubsection{Particle Swarm Optimization}\label{sec:pso}
% introduce PSO; how to apply PSO in this problem
\begin{algorithm}[tb]
    \caption{Particle Swarm Optimisation for Optimal Insertion}
    \label{alg:pso}
    \renewcommand{\algorithmicrequire}{\textbf{Input:}}
    \renewcommand{\algorithmicensure}{\textbf{Output:}}
    \begin{algorithmic}[t]
        \REQUIRE 3D indicator matrix $\bm{M}^{thr}$ from threat volume segmentation, 3D projection cost map $\bm{M}^{bag}$ from bag volume segmentation, inertia weight $w$, cognitive parameter $c_1$, social parameter $c_2$, number of particles $N$.
        \ENSURE Optimal insertion location and orientation $\bm{\Theta}^* = \{x^*,y^*,z^*,\alpha^*,\beta^*,\gamma^*\}$.
        \STATE Set $t=0$ and randomly initialise $\bm{\Theta}_i^0$ and $\bm{V}_i^0$ for $i=1,2,...,N$;
        \WHILE {$t\leq T$}
        \STATE $t \leftarrow t+1$;
        \STATE Calculate $cost[i]$ using Eq. (\ref{eq:obj}) for $i=1,2,...,N$;
        \STATE Find $\bm{\Theta}_i^{best}$ and $\bm{\Theta}^{best}$ according to the calculated $cost$;
        \STATE Update velocities of particles using Eq. (\ref{eq:updateV});
        \STATE Update $\bm{\Theta}^t$ using Eq. (\ref{eq:updateTheta}) for each particle;
        \ENDWHILE
        \STATE Output $\bm{\Theta}^{best}$ as $\bm{\Theta}^*$.
    \end{algorithmic}
\end{algorithm}

To solve the problem defined in Eq. (\ref{eq:obj}), one of the enabling methods is particle swarm optimization (PSO) \cite{kennedy2010particle,pyswarmsJOSS2018}. To make this paper self-contained, we briefly describe the PSO method under our problem setting. The swarm is first initialised with $N$ particles $\{\bm{\Theta}_0^i\}_{i=1}^N$, where each $\Theta^i$ is a vector of six variables. The aim is to find the optimal $\Theta^*$ minimizing the objective function defined in Eq. (\ref{eq:obj}) after $T$ iterations. In the $t$-th iteration, we update the $i$-th particle $\Theta^i_t$ as follows:
\begin{equation}
    \label{eq:updateTheta}
    \bm{\Theta}^t_i = \bm{\Theta}^{t-1}_i + \bm{V}^t_i, i=1,2,...,N,
\end{equation}
where
\begin{equation}
    \label{eq:updateV}
    \bm{V}^t_i = w \bm{V}^{t-1}_i + c_1 r_1 (\bm{\Theta}_i^{best} - \bm{\Theta}_i^{t-1}) + c_2 r_2 (\bm{\Theta}^{best} - \bm{\Theta}_i^{t-1})
\end{equation}
is the velocity of $i$-th particle in the $t$-th iteration; $w$,$c_1$ and $c_2$ are the inertia weight, cognitive and social parameters respectively \cite{shi1998modified,bansal2011inertia}; $r_1$ and $r_2$ are random numbers drawn from a uniform distribution for each particle in each iteration; $\bm{\Theta}^{best}_i$ and $\bm{\Theta}^{best}$ are the best position of $i$-th particle and the best position of the swarm (i.e. all particles) thus far respectively.

After $T$ iterations of Eq. (\ref{eq:updateTheta}-\ref{eq:updateV}), the optimal variables $\bm{\Theta}^* = \{x^*,y^*,z^*,\alpha^*,\beta^*,\gamma^*\}$ can be obtained and are ready to use for threat insertion. The algorithm is shown in Algorithm \ref{alg:pso}.

\subsubsection{Image Blending}\label{sec:blending}
Given the optimal insertion location and orientation, a TIP volume can be generated by inserting the segmented threat object volume into the bag volume after rotating it to the optimal orientation. Recall that the segmented threat volume is denoted as $\bm{I}^{thr}$, it is firstly weighted by the indicator matrix $\bm{M}^{thr}$ such that intensity values of voxels in the \textit{uncertain region} are attenuated. We use the same rotation function $f(\cdot)$ in Eq. (\ref{eq:rotation}) and the optimal angles derived from PSO to rotate the weighted threat volume:
\begin{equation}
    \label{eq:rotateI}
    \tilde{\bm{I}}^{thr} = f(\bm{I}^{thr}\odot \tilde{\bm{M}}^{thr}|\alpha^*,\beta^*,\gamma^*).
\end{equation}
The insertion is an image blending process within 3D imagery in which we modify the values of relevant voxels in the original bag volume $\bm{I}^{bag}$ according to the insertion location and rotated volume of the threat. Different methods can be employed for the purpose of image blending. One simple yet effective method is to add the threat volume matrix $\tilde{\bm{I}}^{thr}$ to the sub-volume matrix of $\bm{I}^{bag}$, where the sub-volume is specified by the optimal position $x^*$,$y^*$ and $y^*$.

\subsection{TIP Quality Score} \label{sec:tqs}
Given a threat volume and a baggage volume, our approach is able to generate a TIP volume which may be of variable comparative realism and plausibility given all possible unconstrained combinations of inputs. On one hand, it is well known that particle swarm optimisation could lead to a local best solution \cite{angeline1998evolutionary}. On the other hand, more importantly, a given baggage volume may just not be suitable for a given randomly selected threat to be inserted as the threat is physically too large (e.g. large firearm threat volume into small handbag target volume). As a result, it is important to have a metric evaluating the quality of a TIP generated volume without manual review. Operationally, this can be used to reject poor quality TIP volumes before the are presented to an operator as part of a TIP based performance evaluation system.

We propose the \textit{TIP quality score} to evaluate the quality of generated TIP volumes. Specifically, we use the cost defined in Eq. (\ref{eq:obj} and normalise it with the volume of inserted threat. The normalised cost value can be easily transformed into a score in the range of $0-100$ using the following equation:
\begin{equation}
    \label{eq:score}
    score = \max(0,\min(100,f(cost)))
\end{equation}
where $f$ is a monotonically  decreasing function which could be selected based on the specific requirement of the TIP quality. In our experiment, we use a simple linear function 
\begin{equation}
    \label{eq:fscore}
    f(cost) = 100-0.01\times cost .
\end{equation}

\subsection{Metal Artefact Generation} \label{sec:mag}
\begin{figure}
\begin{center}
\includegraphics[width=0.8\linewidth]{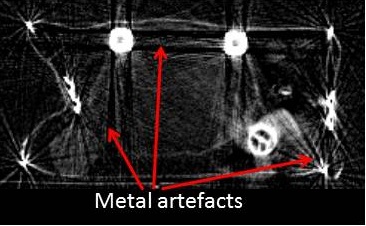}
\end{center}
\caption{CT Metal artefacts in a CT slice of a cluttered baggage.}
\label{fig:mag_example}
\end{figure}

The problem of metal artefacts in X-ray CT images is well studied in medical imaging applications~\cite{ MAcause1, MAR1, MAR2}.  Metal artefacts are caused by the presence of high-density objects in the scan field of view. The origin of metal artefacts has been studied extensively in the literature and several assumptions have been made \cite{mouton2013experimental}. Regardless of the origin of metal artefacts, the effects of these artefacts in the reconstructed CT volumes are the same. Metal artefacts appear as dark and white streaks radiating from the metal objects and spreading across the whole reconstructed CT volumes \cite{megherbi2013radon}. They are more prominent near the metal objects and are a function of scan orientation and the material content (see Figure~\ref{fig:mag_example}).

\begin{figure}[h]
\begin{center}
\includegraphics[width=1\linewidth]{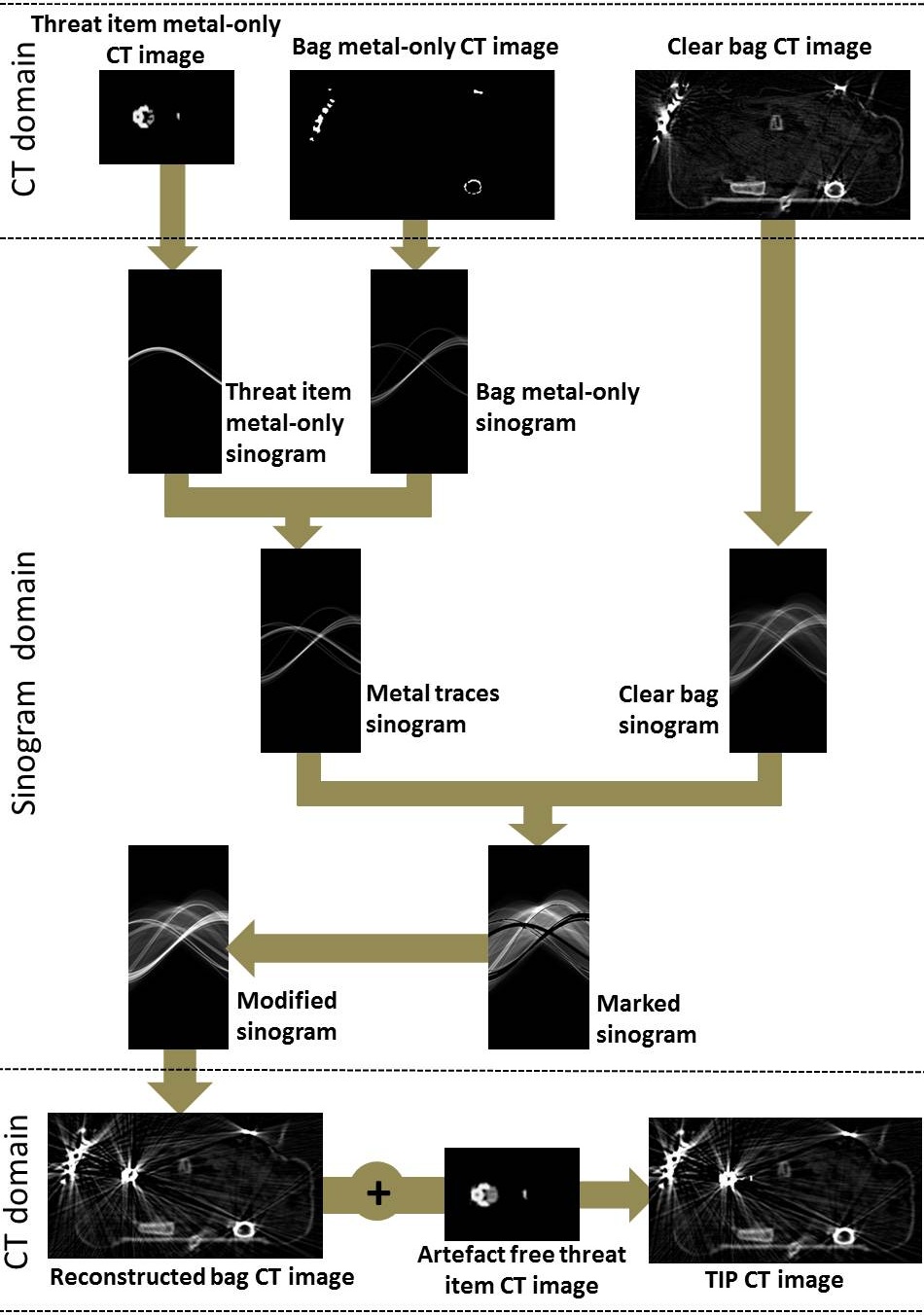}
\end{center}
\caption{Flow chart of our MAG method depicted using 2D CT slices}
\label{fig:mag_pipeline}
\end{figure}

To enhance the plausibility of generated TIP, it is necessary to take into consideration metal artefacts in the TIP process so that the threat objects appear as if they were genuinely located in the scanned bag. 
Our proposed metal artefact generation (MAG) procedure depicted in Figure~\ref{fig:mag_pipeline} is inspired by the established metal artefact reduction (MAR) projection-replacement techniques in medical imaging applications ~\cite{MAR1, MAR2}. 
In a similar vein to these methods, the whole process of MAG is based on a sequence of 2D slices of a 3D CT volume. It starts by mapping the original slices of the benign bag (harmless passenger bag), its metal-only slices and the metal-only slices of the artefact-free threat object to the projection domain via the Radon transform \cite{RT}. The output of this step is known as a sinogram image.
%In order to mimic the CT scanner used during scanning, {\color{red}the CT scanner geometry and the projection parameters used to compute the Radon transform are defined according to the actual CT scanner geometry. This enables our method to include the scan orientation of the bag in the MAG process.} 

The artefact free 3D CT volume of a threat object is obtained by appropriate thresholding to remove artefacts and noise. The metal-only volume of a benign bag and the artefact-free threat object are obtained by segmenting the metal objects in their original CT volume by thresholding using a suitable metal CT intensity threshold. This step exploits the fact that metal objects in CT volumes have higher density compared to other objects. 
Subsequently, the metal traces corresponding to the metal objects of the benign bag and the metal part of the artefact-free threat object are combined in one projection volume. A mask corresponding to all the metal traces is marked in the sinogram of the benign bag CT volume.
In conventional MAR projection-replacement based methods, this mask corresponds to the corrupted area in which projection bins are affected by metallic objects and which need to be replaced by surrogate data. Marking this corrupted area in the Radon domain is equivalent to marking all rays passing through the metallic objects originating from the bag and the threat object in the 3D CT TIP volume (benign bag with the threat object).
CT metal artefacts emerging from the metallic objects spread across these lines. In order to generate metal artefacts in the benign bag CT volume, the projection bins in the marked mask in the benign bag sinogram are thus made inconsistent with their neighbourhood unlike MAR projection-replacement based methods in which the projection bins in this mask are replaced by interpolated data.
The underlying idea behind this is to mimic real CT scanning of a metal object by making the sinogram values corrupted and inconsistent with their neighbourhood if the corresponding X-rays have intersected the metal object. In fact, since metal objects are high-attenuation objects, they heavily attenuate the X-ray beams and consequently, only a few photons reach the scanner detectors. This effect known as photon starvation effect indeed produces corrupted data in the sinogram and gives rise to artefacts in the reconstructed 2D and 3D images.
In order to corrupt the projection bins of the marked mask in the benign bag sinogram, we have used an empirical function as follows:
\begin{equation}
    \label{eq:mag}
    s'_{ij} = (1-q)\times s_{ij} + q\times s_{max}, s_{ij} \in \mathcal{S}_{marked}
\end{equation}
where $s_{ij}$ and $s'_{ij}$ are the benign bag sinogram values within the marked mask before and after being corrupted, respectively; $s_{mask}$ is the maximum value of benign bag sinogram in the marked mask region; $q$ is a hyper-parameter empirically set to $0.2$ in our experiments.
As we will show shortly, by following the above steps, consistent metal artefacts are generated within the bag CT volumes which are a function of the scan orientation of the bag, the material of the bag content and the material of the inserted threat object.
As depicted in Figure~\ref{fig:mag_pipeline}, once the metal artefacts are generated in the Radon space, the resulting modified sinogram is re-projected back into the CT domain. The resulting reconstructed CT volume corresponds to the original benign bag CT volume corrupted by metal artefacts originating from the threat object metal part and the benign bag metal objects.
The final 3D TIP volume is obtained by combining the resulting CT volume with the artefact free threat object CT volume. 
%%%%%%%%%%%%%%%%%%%%%%%%%%%%%%%%%%%% Stop here 7Nov2019 22:23%%%%%%%%%%%%%%%%%%%%%%%

\section{Experiments and Results}\label{sec:experiment}
In this section, experiments are conducted to evaluate the effectiveness of the proposed approach for 3D TIP in baggage CT imagery. In our experiments, the constant value $c$ in Eq. (\ref{eq:bag}) is set to $100$ and the constant value $c'$ in Eq. (\ref{eq:tiphat}) is set to $10$. As a result, voxels with intensity values higher than 410 (i.e. 10/100$\times$4096) will be penalised in the particle swarm optimisation. Values of $\lambda_1$ and $\lambda_2$ in Eq. (\ref{eq:obj}) are empirically set to $0.01$ and $1$ respectively. For each component of the framework, we present some exemplar results in Figures \ref{fig:threatSegRes}-\ref{fig:magRes} for qualitative performance evaluation. We further generate a large number of TIP using real baggage volumes from an airport for quantitative evaluations.
\subsection{Qualitative Evaluations}
\begin{figure}[t]    
    \centering
    \includegraphics[width=0.45\textwidth]{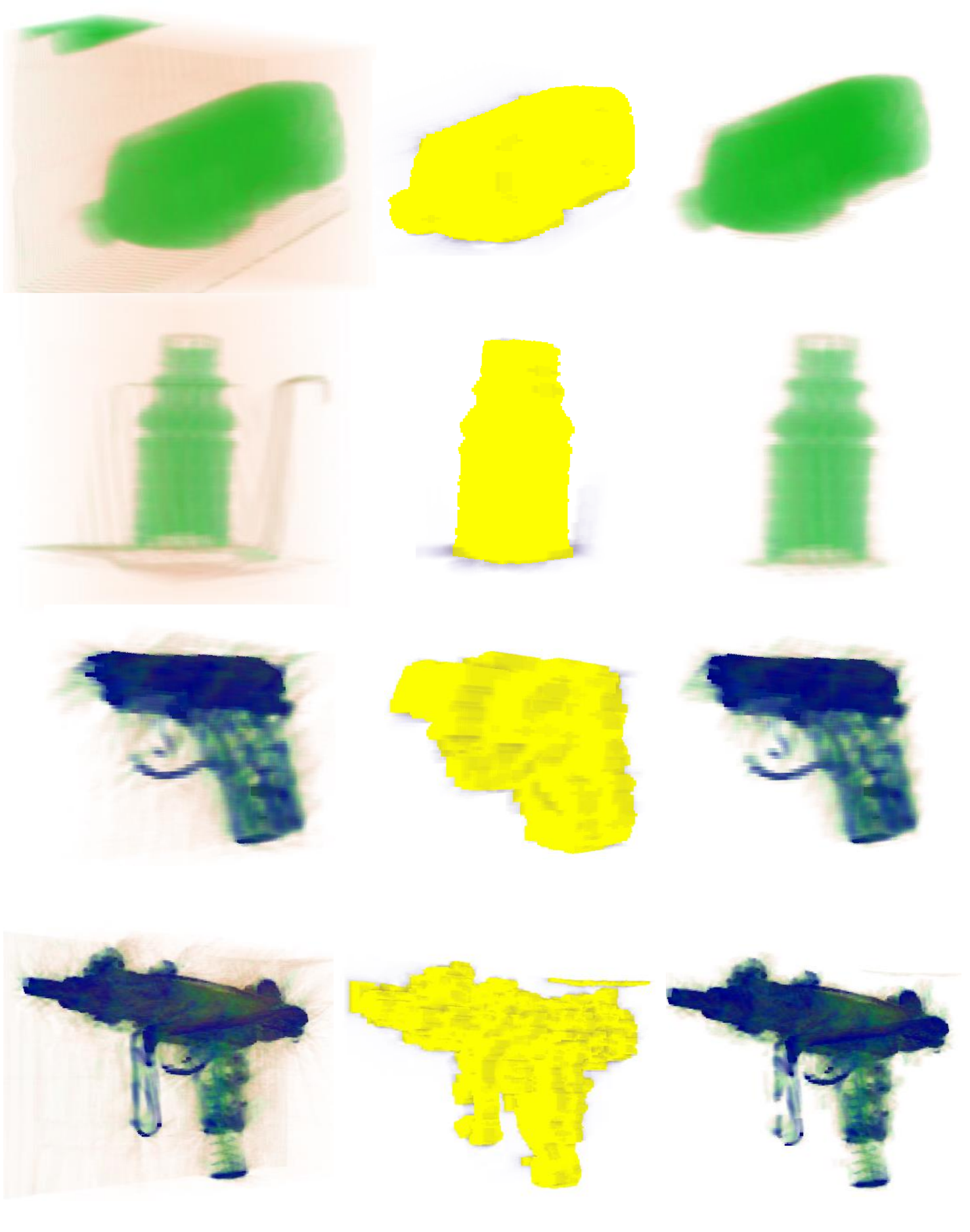}
    \caption{Threat volume segmentation results. Rows from top to bottom: bottle, bottle, handgun and submachine gun. Columns from left to right: original volumes, indicator matrix volumes (defined in Eq. (\ref{eq:threat})) and segmented threat volumes.}
    \label{fig:threatSegRes}
\end{figure}

\begin{figure*}[t]    
    \centering
    \includegraphics[width=\textwidth]{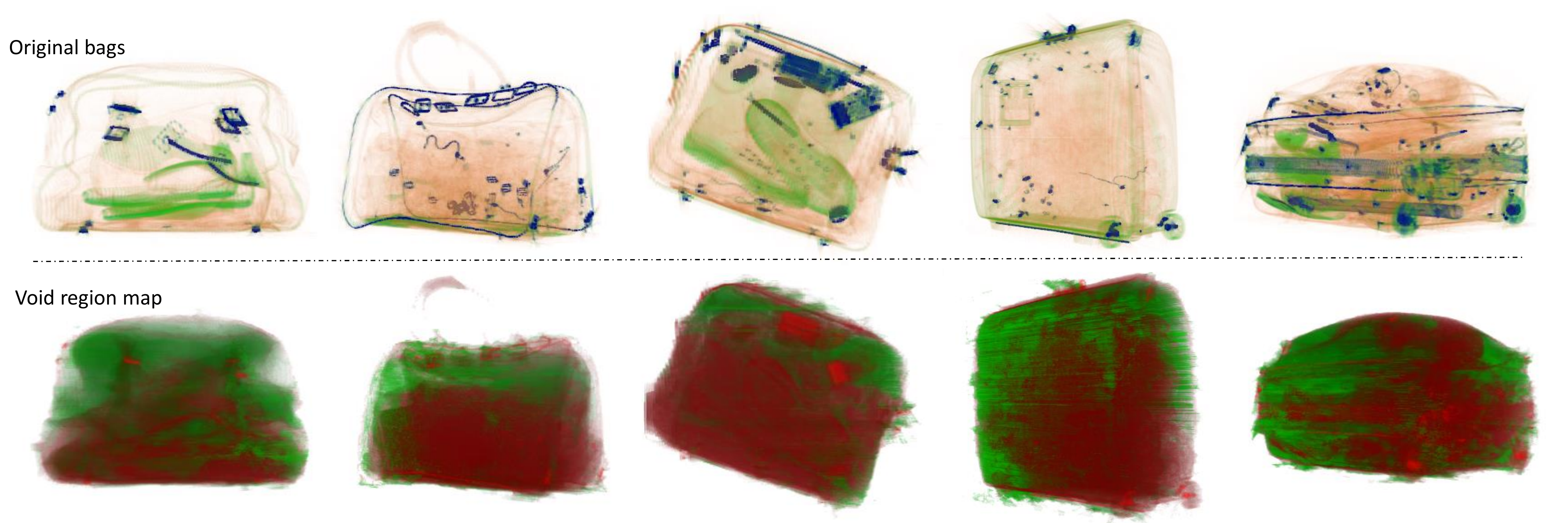}
    \caption{Baggage volume segmentation results. Top row: original baggage CT volumes; bottom row: projection cost volume defined in Eq. (\ref{eq:bag}) with the green colour representing void regions and the red color representing regions having high projection cost.}
    \label{fig:bagSegRes}
\end{figure*}
\begin{figure*}[!t]    
    \centering
    \includegraphics[width=\textwidth]{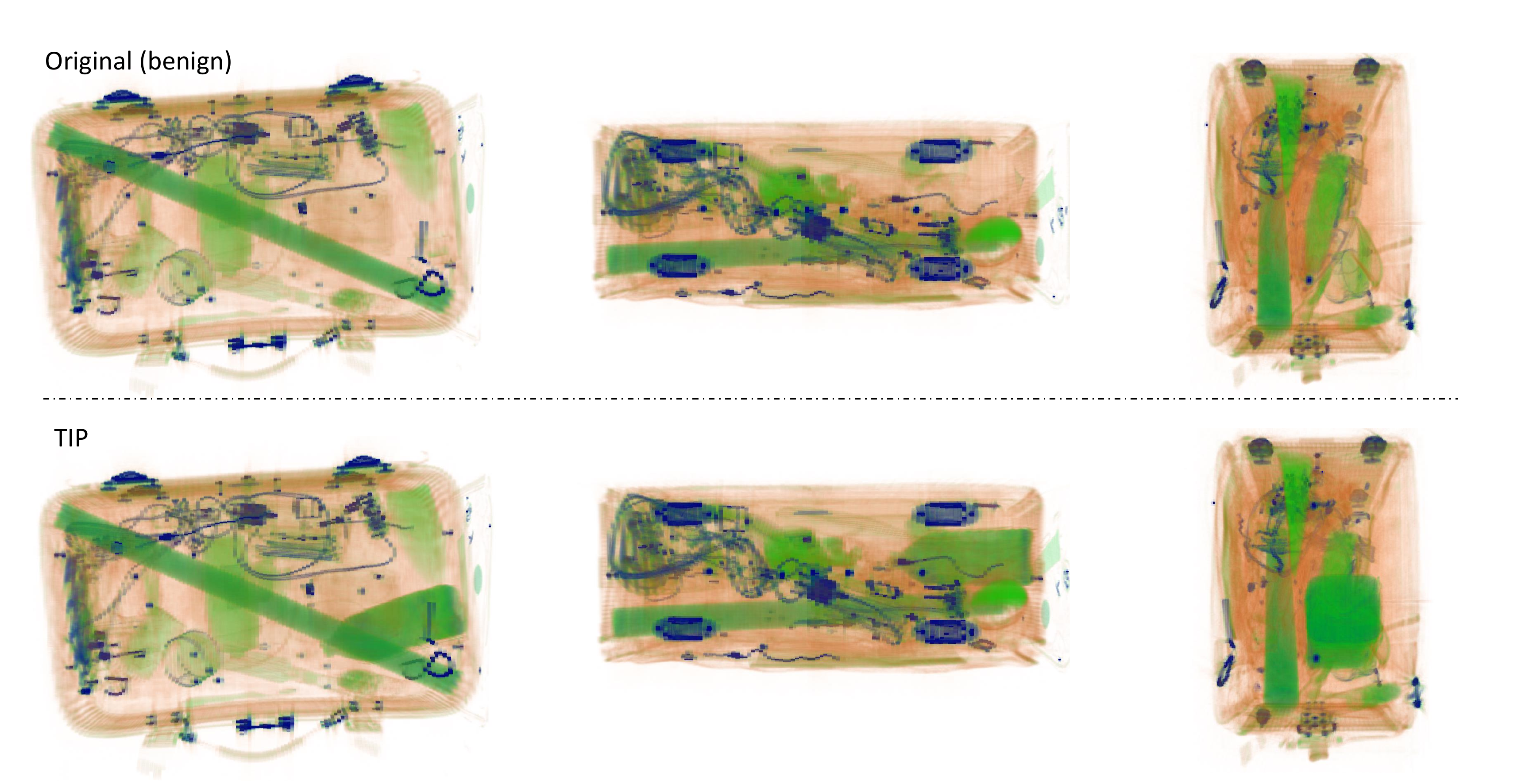}
    \caption{An exemplar threat image projection (TIP) result where a bottle signature is inserted into a suitcase. Three orthogonal views are shown in three columns. Views of the original baggage and the resultant TIP are shown in the top and bottom rows respectively.}
    \label{fig:tipRes1}
\end{figure*}
\begin{figure*}[!t]    
    \centering
    \includegraphics[width=\textwidth]{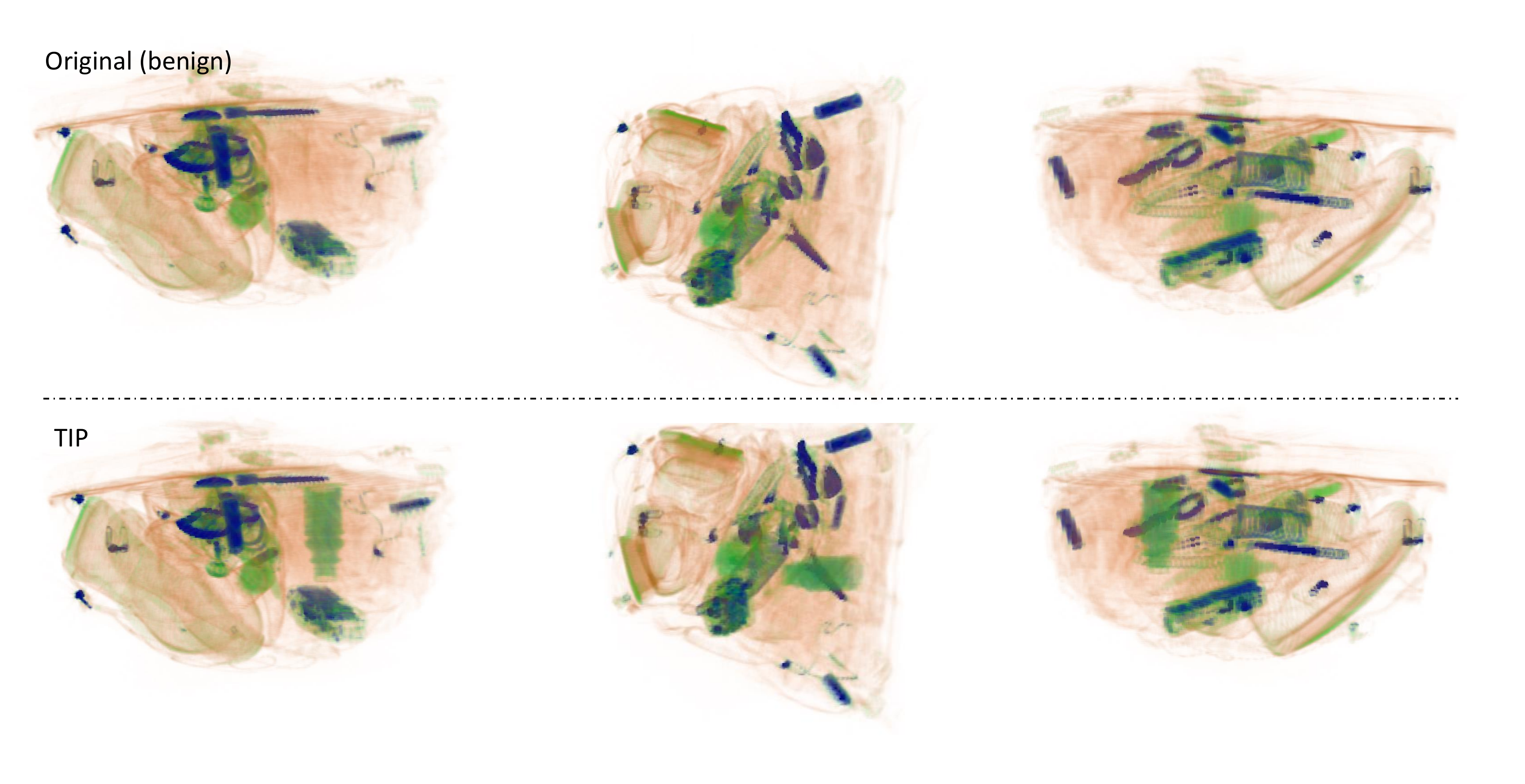}
    \caption{An exemplar threat image projection (TIP) result where a small bottle signature is inserted into a suitcase. Three orthogonal views are shown in three columns. Views of the original baggage and the resultant TIP are shown in the top and bottom rows respectively.}
    \label{fig:tipRes3}
\end{figure*}
\begin{figure*}[!t]    
    \centering
    \includegraphics[width=\textwidth]{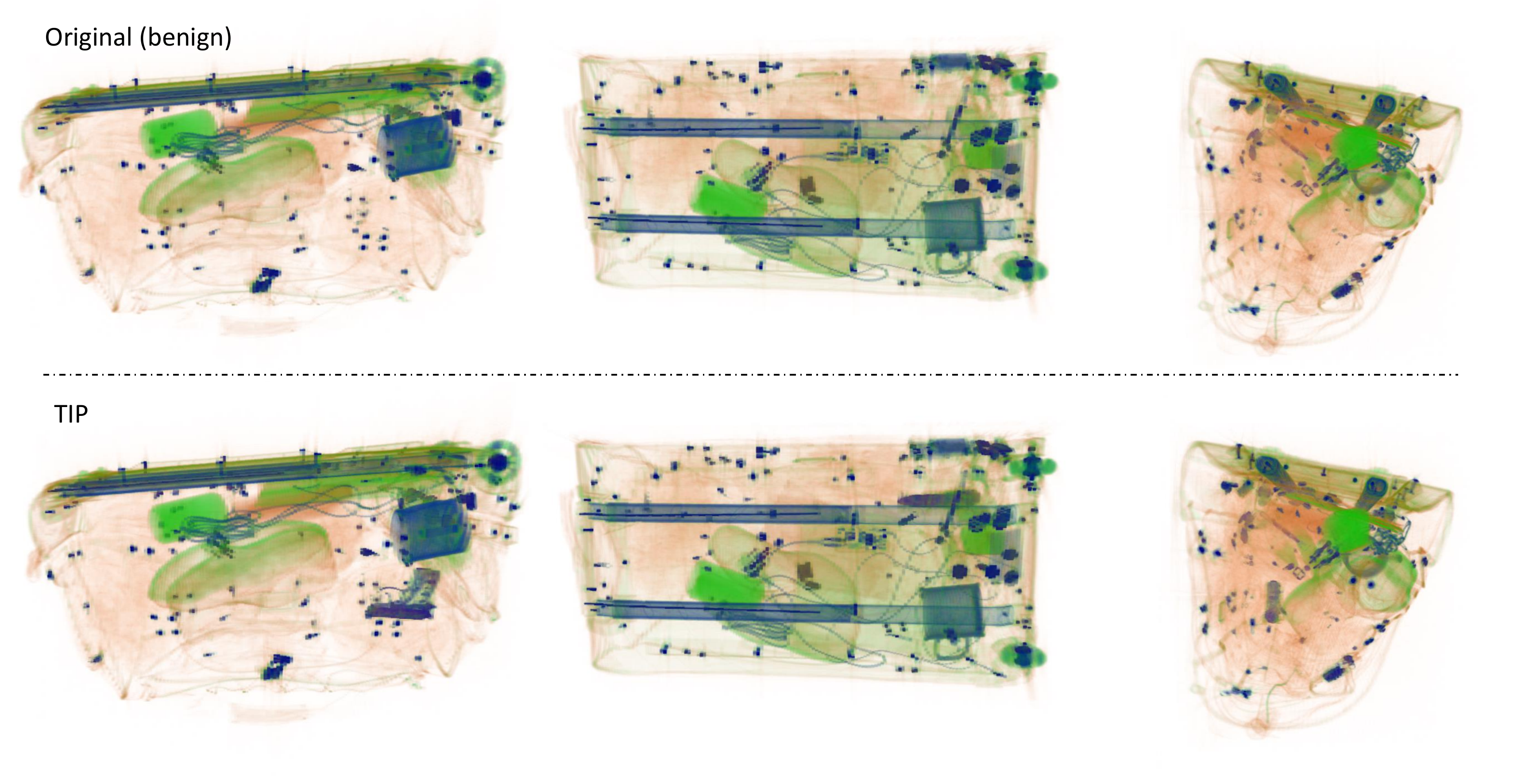}
    \caption{An exemplar threat image projection (TIP) result where a handgun signature is inserted into a suitcase. Three orthogonal views are shown in three columns. Views of the original baggage and the resultant TIP are shown in the top and bottom rows respectively.}
    \label{fig:tipRes-gun1}
\end{figure*}
\begin{figure*}[!t]    
    \centering
    \includegraphics[width=\textwidth]{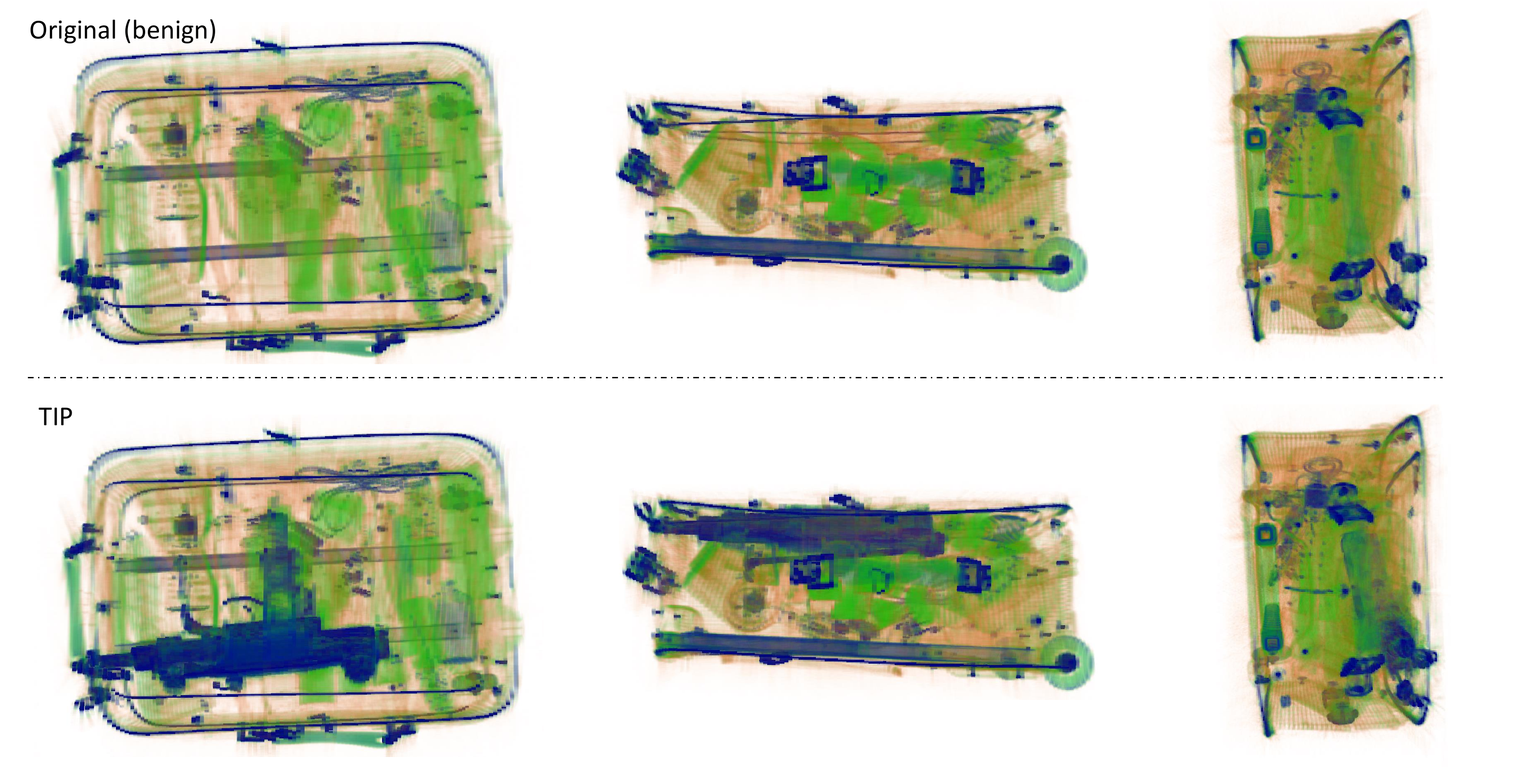}
    \caption{An exemplar threat image projection (TIP) result where a submachine gun signature is inserted into a suitcase. Three orthogonal views are shown in three columns. Views of the original baggage and the resultant TIP are shown in the top and bottom rows respectively.}
    \label{fig:tipRes-miniuzi}
\end{figure*}

\begin{figure}[!]    
    \centering
    \includegraphics[width=0.48\textwidth]{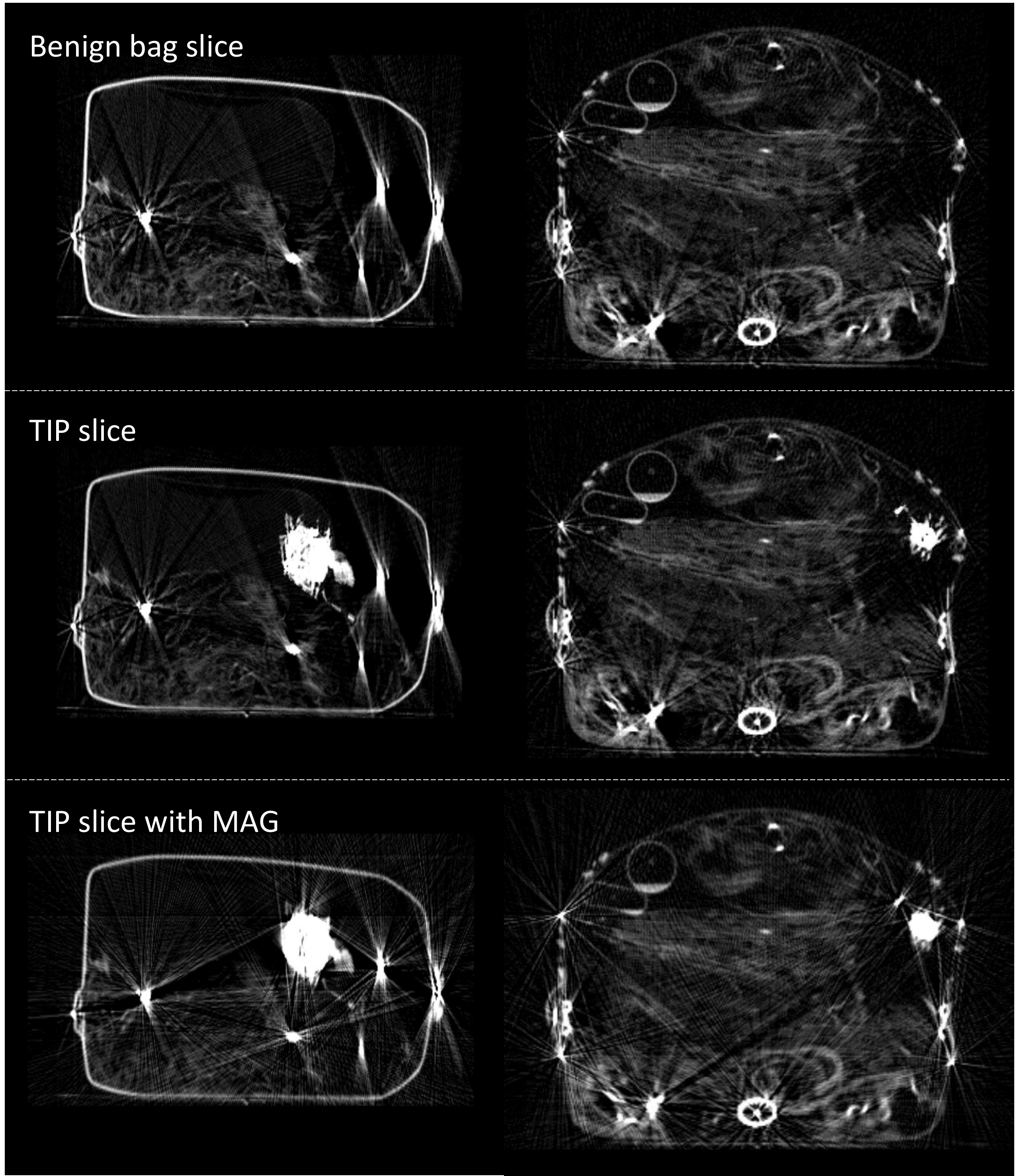}
    \caption{Exemplar results of metal artefact generation (MAG). Two examples are shown in the left and right columns respectively. Rows from top to bottom: slices of benign bag CT volumes; corresponding slices of TIP volumes without MAG; corresponding slices of TIP volumes with MAG.}
    \label{fig:magRes}
\end{figure}

\begin{figure*}[h!]    
    \centering
    \includegraphics[width=\textwidth]{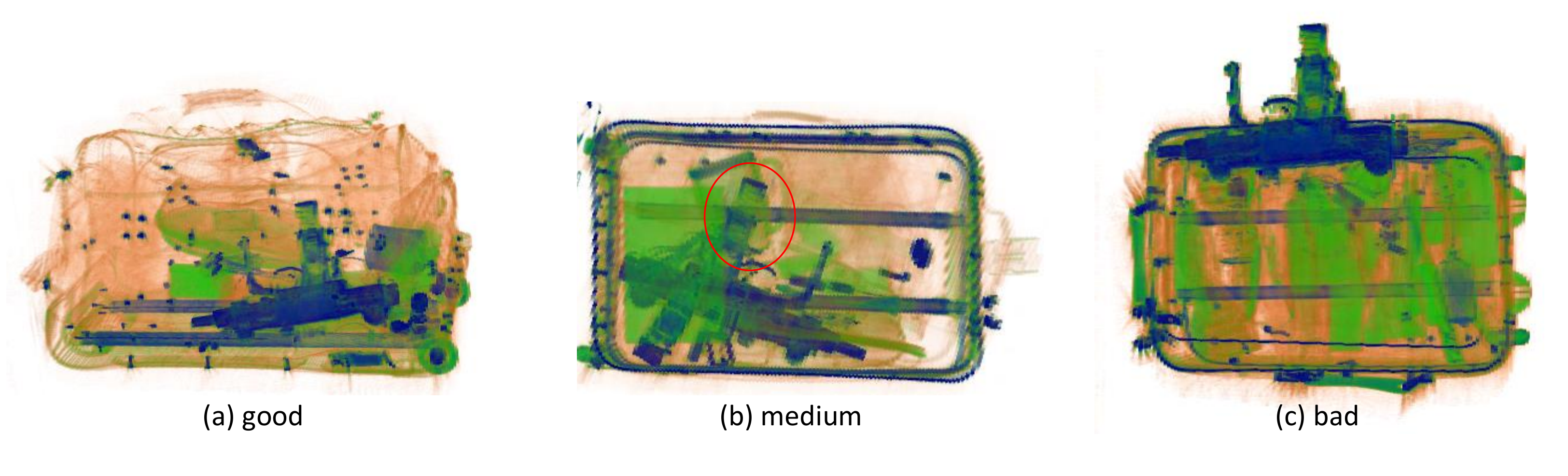}
    \caption{Examples of \textit{good}, \textit{medium} and \textit{bad} TIP volumes: (a) TIP of good quality; (b) TIP of medium quality due to the interception of the magazine into a mug (highlighted in red circle); (c) TIP of bad quality due to the insertion outside the bag.}
    \label{fig:tipexample}
\end{figure*}

Figure \ref{fig:threatSegRes} shows some exemplary threat volume segmentation results using our proposed algorithm. The original threat volumes of two bottles (in the first two rows) and two firearms (handgun in the third row and submachine gun in the fourth row) are displayed in the left column where he background and noises could be observed. The middle column shows the results of our segmentation algorithm indicated by the 3D matrix defined in Eq. (\ref{eq:threat}), where the threat body regions are coloured in yellow, background regions in white and uncertain regions in grey. The segmented threat volumes are shown in the right column from which we can see the background and noises are removed.

Figure \ref{fig:bagSegRes} shows bag volume segmentation results using the proposed algorithm. Five different original bag volumes are displayed in the top row and their corresponding segmentation results are shown in the bottom row. The segmentation results are represented by the 3D projection cost maps defined in Eq. (\ref{eq:bag}), where regions of inner-void, bag-content and background are coloured in green, red and white respectively. The bag segmentation results in Figure \ref{fig:bagSegRes} indicate our algorithm is able to readily locate accurate bag boundaries as well as the void regions inside the bags.

The results of threat volume projection are shown in Figures \ref{fig:tipRes1}-\ref{fig:tipRes-miniuzi}. We visualize the resultant 3D TIP volumes with three orthogonal views in three columns. The first row shows views of the original benign baggage and the TIP results with threat signatures inserted are shown in the second row.
We can see that the threat signatures can be successfully projected into the baggage regardless of the volumes and shapes of the threats and baggage. This attributes to a robust segmentation algorithm for threat and baggage segmentation. Specifically, Figure \ref{fig:tipRes1} shows a TIP result with a bottle projected into a cluttered suitcase. Our approach has been successfully discovered the optimal insertion location and orientation and generated a plausible TIP volume. Figure \ref{fig:tipRes3} shows a TIP result with a small bottle into a backpack. Although the lack of void region in the original backpack, our algorithm projects the bottle signature into a low-intensity region (orange colour). As a result, the inserted bottle looks like being surrounded by organic materials (e.g., clothes) and very realistic. In figure \ref{fig:tipRes-gun1}, the signature of a small handgun is inserted to a baggage and Figure \ref{fig:tipRes-miniuzi} shows the TIP result with a submachine gun signature inserted into a very cluttered suitcase.
In summary, with satisfying results of threat and bag volume segmentation, the particle swarm optimisation algorithm is able to find the optimal position and orientation for the insertion hence plausible TIP could be generated as shown in Figures \ref{fig:tipRes1}-\ref{fig:tipRes-miniuzi}.

To evaluate the performance of the proposed metal artefact generation algorithm, we present two examples in Figure \ref{fig:magRes}. Selected slices of the original bag volumes are shown in the first row. Corresponding slices of the TIP without and with MAG are shown in the second and third rows respectively. We can see that slices with MAG in the third row look more realistic in the region of metal objects due to the generation of artefact streaks.

\subsection{Quantitative Evaluations}\label{eq:quantitative}
We design two experiments to evaluate the proposed TIP approach quantitatively. The first experiment aims to investigate the consistency of TIP quality scores with human evaluations. In Eq. (\ref{eq:score}), a monotonically decreasing function is used in our experiment. %Ten volunteers are recruited to evaluate the quality of TIP images. Before the process, the volunteers are given instructions of what high/low-quality volumes are like. Subsequently, each volunteer is presented with 30 TIP volumes and required to give his/her decision if the TIP volumes are of high/medium/low quality.
We select 150 generated TIP volumes from a large number of candidates and the scores of selected TIP volumes are evenly distributed in the range of $0-100$. We use a 3D CT volume visualisation tool to visually inspect each TIP volume and categorize it into one of three classes (i.e. \textit{good}, \textit{medium} and \textit{bad}) according to their quality. A TIP volume is defined as \textit{good} if the threat signature is perfectly inserted into a void region within the bag volume and visually realistic and plausible. A TIP volume is labelled as \textit{bad} if it is obviously unrealistic, for example, the threat signature is inserted outside the bag or intercepted by other items in the bag. A TIP volume of \textit{medium} quality is not perfect but the flaw can only be spotted by careful inspection after the considerable time ($>$ 2 minutes). As a result, there are 102, 37 and 11 TIP volumes labeled as \textit{good}, \textit{medium} and \textit{bad} respectively. It is noteworthy this ratio is not a performance reflection of our TIP approach since we deliberately select low score TIP to look into the relationship between TIP scores and TIP qualities in this experiment. We calculate the mean and standard deviation of the scores for the TIP volumes falling in each class and the results are $82.8 \pm 23.2$, $69.8\pm 36.1$ and $45.7\pm 48.1$, respectively. In general, the proposed TIP quality scores are consistent with human evaluation results in terms of mean values. On the other hand, however, we also can see large standard deviations of the quality scores for all three classes. It is indicated that the TIP quality score is not perfectly reliable for TIP quality evaluation. As a result, we need to set a high score threshold to reject TIP of bad quality in practice which unavoidably will also falsely reject some good ones.

In the second experiment, we randomly select 100 generated TIP volumes and manually label each of them as the class of \textit{good}, \textit{medium} or \textit{bad} based on visual inspection. As a result, 92\% of the TIP volumes are good, 6\% are medium and only 2\% of them are bad. These results demonstrate that our proposed TIP approach is able to generate TIP volumes of good quality with a very high plausibility and realism acceptance rate.

\section{Discussion}\label{sec:discussion}
In this section, we discuss the limitations of our proposed approach and potential solutions to addressing them in future work. Specifically, we discuss three aspects: preconditions, hyper-parameters and failure cases.

The behaviours of our approach rely heavily on values of many hyper-parameters such as threshold values in volume segmentation and insertion location optimisation. Although the approach is generally robust to most parameters hence performs well as illustrated without exhaustive parameter tuning, there is a subset we need to take special care with when applying this approach to CT volumes captured with different scanners. This is due to the considerable variability of voxel value ranges and noise levels of CT machines from different manufacturers or even different models from the same manufacturer.

One parameter we need to adjust for specific scanners is the threshold value for binarization in the first step of void determination (c.f. Figure \ref{fig:bagSegAlg}). This threshold determines the accuracy of bag boundary. We are aware that if this threshold value is too small, the noise surrounding the bag would be mistakenly treated as part of the bag. As a result, a region outside the bag could be potentially treated as inner void and be the place where threats are inserted in. This would make the resultant TIP obviously unrealistic. One way to addressing this issue is to set this threshold value higher, which, however, could mistakenly remove voxels of bag boundaries since the materials of bag surfaces usually have low intensities in CT volumes. Fortunately, this kind of TIP artefact is more acceptable compared with the former one (i.e. inserting threats outside a bag). We, therefore, suggest a great value rather than a small one for the threshold of binarization in bag volume segmentation.

\section{Conclusion}\label{sec:conclusion}
In this paper, we propose an approach to 3D threat image projection for X-ray CT volumes. Qualitative and quantitative evaluations prove that our TIP approach is able to generate realistic and plausible 3D baggage CT volumes containing fictional threat signatures which can be widely used for training baggage screening operators or automatic threat detection models extending of \cite{megherbi13segmentation}. In our future work, we will investigate how the deep convolutional neural networks can be employed and benefit the performance of bag and threat image segmentation in our TIP approach. On the other hand, it will be more interesting to investigate how the generated TIP volumes could benefit the training of automatic threat detection models as one type of data augmentation strategy \cite{bhowmik2019good}.

\bibliographystyle{IEEEtran}
\bibliography{ref}
\end{document}